\documentclass[sigconf]{acmart}

\usepackage{graphicx}
\usepackage{subcaption} 
\usepackage{hyperref}
\usepackage{xcolor} 
\usepackage{multirow}

\newtheorem{remark}{Remark}

\AtBeginDocument{%
  }

\setcopyright{acmlicensed}

\copyrightyear{2026}
\acmYear{2026}
\setcopyright{cc}
\setcctype{by}
\acmConference[CIKM '26]{Proceedings of the 35th ACM International Conference on Information and Knowledge Management}{November 07--11, 2026}{Rome, Italy}
\acmBooktitle{Proceedings of the 35th ACM International Conference on Information and Knowledge Management (CIKM '26), November 07--11, 2026, Rome, Italy}
\acmDOI{10.1145/3799682.3840053}
\acmISBN{979-8-4007-2539-5/2026/11}

\begin{document}

\title{Stitch the Fragments: One-Shot Hierarchical Federated Clustering}

\author{Shenghong Cai}
\affiliation{%
  \institution{School of Computer Science and Technology, Guangdong University of Technology, Guangzhou, China}
  \country{}}
\email{3121005074@mail2.gdut.edu.cn}

\author{Zihua Yang}
\affiliation{%
  \institution{School of Computer Science and Technology, Guangdong University of Technology, Guangzhou, China}
  \country{}}
\email{3122004153@mail2.gdut.edu.cn}

\author{Yuzhu Ji}
\affiliation{%
  \institution{School of Computer Science and Technology, Guangdong University of Technology, Guangzhou, China}
  \country{}}
\email{yuzhu.ji@gdut.edu.cn}

\author{Yiqun Zhang}
\correspondingauthor
\affiliation{%
  \institution{School of Computer Science and Technology, Guangdong University of Technology, Guangzhou, China}
  \country{}}
\email{yqzhang@gdut.edu.cn}

\author{Mengke Li}
\affiliation{%
  \institution{College of Computer Science and Software Engineering, Shenzhen University, Shenzhen, China}
  \country{}}
\email{mengkeli@szu.edu.cn}

\author{Yang Lu}
\affiliation{%
  \institution{School of Informatics, Xiamen University, Xiamen, China}
  \country{}}
\email{luyang@xmu.edu.cn}

\author{Yiu-Ming Cheung}
\affiliation{%
  \institution{Department of Computer Science, Hong Kong Baptist University, Hong Kong SAR, China}
  \country{}}
\email{ymc@comp.hkbu.edu.hk}

\renewcommand{\shortauthors}{Shenghong Cai et al.}
%% No italics, no superscripts, not anonymous
%% Use footnote or author note to identify equal contribution, shared contribution, and/or contact author info

\begin{abstract}

Federated Clustering (FC) faces a critical bottleneck in real-world scenarios, i.e., global clusters are rarely intact, often fragmenting into incomplete, multi-granular unlabeled ``clusterlets'' distributed across Non-IID clients. Although hierarchical clustering is theoretically well-suited to model such nested distributions, its recursive nature strictly relies on multi-round communication, introducing prohibitive computational overhead and severe privacy vulnerabilities. This paper, therefore, proposes a novel one-shot hierarchical federated clustering framework designed to seamlessly ``stitch'' the fragmented local clusterlets into a holistic global distribution. Our approach enables clients to perform autonomous fine-grained distribution exploration, uploading prototype-level knowledge via a dynamic parameter-interleaving mechanism to scramble transmission trajectories, which effectively prevents the server from tracing individual client data distributions. Subsequently, a multi-granular learning mechanism at the server fuses these granularly inconsistent local clusterlets, reconstructing a coherent global hierarchy for ultimate clustering. Extensive experiments on real benchmark datasets illustrate the superiority of the proposed approach, which effectively bridges the granularity gap among heterogeneous clients while minimizing privacy exposure risks via anonymized informative one-shot communication. Source code is accessible \href{https://github.com/hongdream/Fed-HIRE}{\underbar{\textcolor{blue}{here}}}.

\end{abstract}




\maketitle

\section{Introduction}
Federated Clustering (FC) has emerged as a key area of interest in federated learning~\cite{10.1145/3298981,9084352,Li2023FLSurvey}, aiming to partition global data by utilizing client-side resources~\cite{10.1561/2200000083} while preserving privacy~\cite{Wei2020DPFL,Yin2021PrivacySurvey,11397203}.
It offers a general solution for distributed scenarios that lack label guidance, such as collaborative disease discovery~\cite{LI2020106854}, user profiling~\cite{9060868}, and recommendation~\cite{He2024CoFedRec}.
However, due to the non-IID nature~\cite{ZHU2021371,Li2022NonIID,Qin2023FedAPEN} and the resulting distributional mismatch~\cite{Liu2025MindGap}, the absence of ground-truth labels makes aligning divergent local cluster distributions particularly challenging, consistent with observations in clustered federated learning~\cite{Ghosh2020CFL,Sattler2021CFL}.
This issue is further exacerbated when local clusters exist at varying granularities~\cite{10631083}, as misalignment across granular levels can severely degrade the overall clustering performance.

Most existing FC methods allow clients to learn local cluster distributions~\cite{9516933,9499980,YAN2024121203} and then share communication-efficient summaries~\cite{9933478} or privacy-preserving knowledge~\cite{Li2024DPFedC} with the server for global aggregation.
Prototype-based communication has also been explored in federated unsupervised representation learning~\cite{Zhang2024Prototype}.
For instance, FedSC~\cite{NEURIPS2023_b6cd2650}, FFCM~\cite{stallmann2022towards}, and AFCL~\cite{Zhang_Zhang_Lu_Li_Chen_Cheung_2025} require multiple communication rounds to iteratively refine global clusters, incurring substantial overhead and privacy risks. 
Although one-shot alternatives, e.g., Fed-SC~\cite{10184550} and GOLD~\cite{11328086}, mitigate communication overhead, recent FC studies have also explored the unknown number of clusters and hierarchical aggregation~\cite{Zou2024UnknownK,Zhang2026FedHC}. Yet existing hierarchical solutions, including centralized streaming HC~\cite{Han2022StreamingHC}, do not address one-shot reconstruction from fragmented, granularity-inconsistent clusterlets.

\begin{figure*}[!t]
    \centering
    \includegraphics[width=0.99\linewidth]{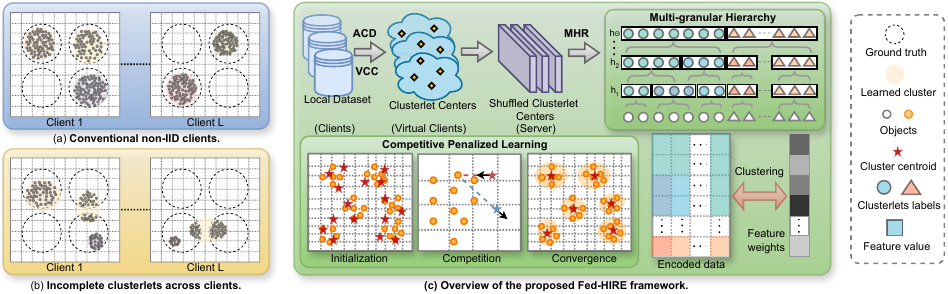}
    \caption{
    (a) The typical non-IID assumption expects each client's data to represent the complete distribution of its clusters.
(b) In contrast, real-world clusters are fragmented across clients into multiple clusterlets.
(c) Overview of Fed-HIRE. Each client performs Adaptive Clusterlet Discovery (ACD) to identify local clusterlets, shuffles their centroids via virtual clients, and uploads anonymized information to the server. The server performs Multi-Granular Hierarchy Reconstruction (MHR) to organize the received clusterlets into a global hierarchy, which is embedded into encoded representations for global clustering.
    }
    \label{fig:intro_diagram}
\end{figure*}

Notably, current approaches typically rely on an ideal non-IID assumption~\cite{pmlr-v139-dennis21a}, expecting each client to capture complete local cluster distributions (see Fig.~\ref{fig:intro_diagram}(a)). In reality, however, global clusters are often fragmented into incomplete, heterogeneous \textit{clusterlets} distributed across clients (see Fig.~\ref{fig:intro_diagram}(b)). This multi-granular inconsistency causes local clients to misidentify these fragments as distinct clusters, obscuring their shared global identities. Consequently, this fragmentation poses three critical challenges: 1) the presence of incomplete clusters complicates the exploration of client-specific distributions, 2) the misalignment of granularity levels across clients impedes hierarchical structure aggregation at the server, and 3) the necessity of transmitting fine-grained structural details for global reconstruction fundamentally conflicts with the privacy-preserving mandate of federated learning~\cite{LiLuo2024PrivacyFC,Chen2024PrivCrFL}.

To address these challenges, we propose a novel one-shot Federated HIerarchical Representation Enhancement framework, termed \textbf{Fed-HIRE}, to seamlessly ``stitch'' the structural fragments into a holistic global distribution. First, each client employs Adaptive Clusterlet Discovery (ACD) to estimate the number of local clusterlets and capture fine-grained local distributions. To safeguard privacy, a streaming Virtual Client Confusion (VCC) mechanism shuffles transmission trajectories and reorganizes local centroids into anonymized uploads. Subsequently, the server performs Multi-Granular Hierarchy Reconstruction (MHR) to integrate the received centroids and organize the global cluster distribution into a coherent hierarchy. Finally, the learned multi-granular structural information is embedded into data-enhanced representations to yield the final partition. The main contributions are as follows:

\begin{itemize}
\item A novel one-shot FC framework is introduced to tackle the challenging unsupervised distribution aggregation problem under highly fragmented scenarios. Uniquely, it leverages the reconstructed multi-granular hierarchy as a versatile representation basis, embedding multi-granular structural knowledge to flexibly enhance the global clustering tasks.
\item A prototype-driven self-organizing learning mechanism is developed to dynamically fuse multi-granular local clusterlets at the server. It successfully reconstructs a holistic global hierarchy, effectively overcoming the information bottleneck imposed by the limited distributional details available from one-shot, anonymized prototypes.
\item An anonymized communication mechanism is proposed to scramble transmission trajectories by dynamically interleaving uploaded parameters. More importantly, this mechanism fundamentally minimizes privacy exposure risks without degrading the structural utility, establishing a highly secure paradigm for unsupervised knowledge aggregation.
\end{itemize}

\section{Proposed Method}

The FC problem is formulated as follows: considering a federated network with $L$ distributed clients, the global dataset is distributed across these clients, where the $l$-th client possesses a private dataset $\mathbf{X}^{(l)} = \{ \mathbf{x}_i^{(l)} \}_{i=1}^{n^{(l)}}$ consisting of $n^{(l)}$ objects. Each object has $d$ features, and the total number of objects across the network is $n = \sum_{l=1}^{L} n^{(l)}$. The objective of federated clustering is to partition these $n$ distributed objects into $k^*$ global clusters, denoted as $\mathcal{C} = \{ C_j \}_{j=1}^{k^*}$, with the corresponding cluster centers denoted as $\{ \mathbf{c}_j \}_{j=1}^{k^*}$. Specifically, we maximize the total intra-cluster similarity, formulated as:

\begin{equation}
\label{eq:objective}
P(\mathcal{Q}, \mathcal{C}) = \sum_{l=1}^{L} \sum_{i=1}^{n^{(l)}} \sum_{j=1}^{k^*} q_{ij}^{(l)} s\left(\mathbf{x}_i^{(l)}, \mathbf{c}_j\right),
\end{equation}
where $s(\cdot, \cdot)$ measures the similarity between a data object and a global cluster center, defined as:
\begin{equation}
\label{eq:similarity}
s\left(\mathbf{x}_i^{(l)}, \mathbf{c}_j\right) = \frac{1}{\|\mathbf{x}_i^{(l)} - \mathbf{c}_j\|^2_2 + 1}.
\end{equation}
Here, $\mathcal{Q} = \{ \mathbf{Q}^{(l)} \}_{l=1}^{L}$ represents the object-cluster assignments across all clients, where the binary entry $q_{ij}^{(l)} \in \{ 0, 1 \}$ indicates whether the $i$-th object on the $l$-th client is assigned to the $j$-th global cluster, satisfying $\sum_{j=1}^{k^*} q_{ij}^{(l)} = 1$.

In the Fed-HIRE framework, local clients first execute the Adaptive Clusterlet Discovery (ACD) mechanism to explore clusterlets. Subsequently, the results are shuffled and anonymized through the Virtual Client Confusion (VCC) mechanism before being uploaded to the central server. Upon receiving these scrambled centroid matrices, the server deploys the Multi-Granular Hierarchy Reconstruction (MHR) algorithm to uncover a coherent global hierarchy. Finally, Fed-HIRE embeds these multi-granular hierarchies into data-enhanced representations to facilitate highly informative global clustering. An architectural overview of the full Fed-HIRE is illustrated in Fig.~\ref{fig:intro_diagram}(c).

\noindent\textit{\textbf{Client-Side Exploration: Adaptive Clusterlet Discovery (ACD).}}
To explore fine-grained data distributions while adaptively estimating the appropriate number of clusterlets, on the client side, we introduce the Adaptive Clusterlet Discovery (ACD) algorithm, which initializes a large number of candidate clusterlets $\mathcal{C}^{(l)} = \{ C^{(l)}_j \}_{j=1}^{k_0}$ with a sufficiently large $k_0$. 
These candidate clusterlets compete with each other during optimization, which effectively captures the fine-grained data distribution by preserving prominent clusterlets while gradually eliminating redundant ones.

For the $l$-th client, we introduce a clusterlet weight set $W = \{ w_j \}_{j=1}^{k_0} $, where $w_j$ denotes the importance of the candidate clusterlet $C^{(l)}_j$. For each object $x^{(l)}_i$, its winning clusterlet $C_v^{(l)}$ is determined using the similarity in Eq.~\eqref{eq:similarity}:
\begin{equation}
\label{eq:v}
v = \arg \max_{1 \le j \le k_0} {\left [ \gamma_j w_j s\left(\mathbf{x}_i^{(l)}, \mathbf{c}_j^{(l)}\right) \right]},
\end{equation}
where $\gamma_j$ is the winning-frequency balancing factor of $C_j^{(l)}$, computed as:
\begin{equation}
\label{eq:gamma}
\gamma_j=1-\frac{g_j}{ {\textstyle \sum_{t=1}^{k_0}g_t} }.
\end{equation}
Here, $g_j$ records the winning frequency of $C^{(l)}_j$ during optimization, which is updated via $g_v = g_v + 1$.
% \begin{equation}
% \label{eq:g}
% g_v = g_v + 1.
% \end{equation}
The nearest rival clusterlet $C_r^{(l)}$ relative to $C_v^{(l)}$ is identified as:
\begin{equation}
\label{eq:r}
r = \arg \max_{1 \le j\le k_0, j \ne v} {\left [ \gamma_j w_j s\left(\mathbf{x}_i^{(l)}, \mathbf{c}_j^{(l)}\right) \right]}.
\end{equation}
%\begin{remark}
%\textbf{Fair Clusterlet Competition:} The parameter $\gamma_j$ mitigates the dominance of frequently winning clusterlets during optimization by dynamically reducing their relative winning probabilities. This formulation ensures that different initialized clusterlets are fairly evaluated, which facilitates the effective elimination of redundant clusters. Concurrently, it can avoid the effect that some initialized clusterlets located in marginal positions will immediately become the ``dead units'' without learning updates. 
%\end{remark}

The clusterlet weight $w_j$ in Eq.~\eqref{eq:v} needs to be dynamically updated to reflect the importance of each clusterlet. To constrain the clusterlet weights within the interval $[0, 1]$, we introduce an auxiliary variable $\tau_j$ to update $w_j$. Specifically, we reward the winning clusterlet $C^{(l)}_v$ by $\tau_v = \tau_v + \eta$,
and penalize the nearest rival clusterlet $C^{(l)}_r$ by decreasing $\tau_r$:
\begin{equation}
\label{eq:r_weight}
\tau_r = \tau_r - \eta \cdot s\left(\mathbf{x}_i^{(l)}, \mathbf{c}_r^{(l)}\right),
\end{equation}
where $\eta$ denotes the learning rate. Through this competitive mechanism, the winning clusterlet $C^{(l)}_v$ is reinforced by a step size of $\eta$, enhancing its likelihood of being preserved as a dominant structure. Conversely, the rival clusterlet $C^{(l)}_r$ is adaptively penalized based on its similarity to the object. Subsequently, the sigmoid function is employed to map $\tau_j$ into $w_j$, formulated as:
\begin{equation}
\label{eq:client_weight}
w_j = \frac{1}{1+e^{-10(\tau_j + 5)}}.
\end{equation}

When the clusterlet weight $w_j$ approaches 0, the corresponding clusterlet is eliminated, and its associated data objects are reassigned to the remaining active clusterlets. In this way, the $l$-th client ultimately obtains $k^{(l)}$ clusterlets. 

\noindent\textit{\textbf{Privacy-Preserving Transmission: Virtual Client Confusion (VCC).}}
To safeguard privacy against potential server-side intersection attacks and trajectory reconstruction, we introduce the Virtual Client Confusion (VCC) mechanism. Operating as an anonymization layer, VCC dynamically maps and routes local information through a sequence of Virtual Clients (VCs). This process effectively scrambles physical transmission trajectories and disrupts the structural composition of the uploaded knowledge from each client, rendering individual data distributions untraceable by the server.

Let $\mathcal{C}^{(l)} = \{ \mathbf{c}^{(l)}_j \}_{j=1}^{k^{(l)}}$ denote the set of learned centroids of the $l$-th client, where $k^{(l)}$ is the estimated number of local clusterlets. Upon visiting a physical client, each VC dynamically samples a random subset of clusterlet centroids with a randomized cardinality bounded within $[1, \delta]$ from the respective local centroid matrix, where $\delta$ denotes a predefined upper bound on the number of cluster centroids. Once a VC completes a full traversal of the clients, it acts as an anonymous uploading node and is staged in a transmission queue. This execution persists until all clusterlet centroids from all physical clients have been uploaded. Assuming a total of $M$ VCs are instantiated, the unified structured data transmitted to the central server is formalized as $\mathcal{C} = \{ \mathbf{C}^{(m)} \}_{m=1}^M$.

\noindent\textit{\textbf{Server-Side Aggregation: Multi-Granular Hierarchy Reconstruction (MHR).}}
At this stage, the server receives the clusterlet centroids from all VCs. By stacking all received sets $\mathcal{C}^{(m)}$, we obtain a global centroid matrix $\mathbf{C} \in \mathbb{R}^{n^{(s)} \times d}$, where $n^{(s)}$ is the total number of uploaded clusterlet centroids. To uncover the global cluster structure, we introduce the Multi-Granular Hierarchy Reconstruction (MHR) algorithm, which automatically uncovers multi-granular global distributions and constructs a structured cluster hierarchy.

To establish the foundation of this hierarchy, MHR starts from a significantly large initial cluster number $k_0$, applying the competitive penalized learning mechanism (Eqs.~\eqref{eq:v}-\eqref{eq:client_weight}) to partition $\mathbf{C}$ into $k_1$ clusters, yielding the finest-grained object-cluster assignment matrix $\mathbf{Q}_1$. To explore coarser distributions, this optimization process is recursively executed. Each subsequent round inherits the converged cluster count from the preceding stage (e.g., $k_1$) and resets all internal learning factors. The recursion terminates when no further meaningful coarser structures can be identified. Consequently, the server systematically uncovers the global distribution from the finest to the coarsest granularity levels, yielding a sequence of monotonically decreasing cluster counts (i.e., $k_1 > k_2 > \cdots > k_{\Theta}$). By aggregating the outcomes across all stages, MHR constructs a comprehensive hierarchical structure $H$, formally defined as a collection of multi-granular object-cluster assignment matrices alongside their corresponding cluster counts:
\begin{equation}
\label{eq:hierarchy}
H = \{ (\mathbf{Q_\theta}, k_\theta) \mid \mathbf{Q_\theta} \in \{0, 1\}^{n^{(s)} \times k_\theta}, 1 \le \theta \le \Theta \},
\end{equation}
where $\theta$ denotes the $\theta$-th granularity level of the hierarchy, and $\Theta$ represents the total number of hierarchical levels.

\begin{remark}
\textbf{Reinitialization for Structural Diversity:} At each stage of MHR, only the cluster count $k$ learned from the preceding stage is inherited to initialize the next stage, while the corresponding cluster centroids are randomly re-initialized. This repeated re-initialization facilitates a comprehensive exploration of global distributions, allowing each layer to effectively capture the underlying data distribution and enabling the construction of a meaningful hierarchy. Conversely, directly inheriting the converged centroids from previous stages introduces strong structural dependencies, reducing structural diversity across different granularities.
\end{remark}

\noindent\textit{\textbf{Representation Enhancement for Holistic Clustering.}}
To obtain the final clustering partition with respect to the target number of global clusters $k^*$ based on the multi-granular hierarchy $H$, Fed-HIRE embeds the object-cluster assignments from each layer $\mathbf{Q}_\theta \ (1 \le \theta \le \Theta)$ into a data-enhanced representation $\mathbf{X}^{(s)} \in \mathbb{R}^{n^{(s)} \times \Theta}$. Specifically, for the $n^{(s)} \times k_\theta$ matrix $\mathbf{Q}_\theta$ at the $\theta$-th hierarchical level, the corresponding encoded information is mapped into the $\theta$-th column of $\mathbf{X}^{(s)}$ via:
\begin{equation}
\label{eq:encode}
x_{i\theta}^{(s)}
=
\arg\max_{1\le j\le k_\theta}
q_{ij}^{(\theta)},
\qquad
1\le i\le n^{(s)},\quad
1\le\theta\le\Theta.
\end{equation}

As the $\theta$-th column of $\mathbf{X}^{(s)}$ captures the cluster structural information at the $\theta$-th granularity level, the contribution of each granularity to the final partition varies. To accommodate these differing contributions, Fed-HIRE performs clustering on $\mathbf{X}^{(s)}$ by incorporating a feature weight vector $\mathbf{f} = [f_1, f_2, \dots, f_{\Theta}]^T$. We use $\mathbb{I}[\beta]$ to denote the indicator function, which equals 1 if proposition $\beta$ is true and 0 otherwise. Following the global optimization objective in Eq.~\eqref{eq:objective}, the cluster assignment for each data object $\mathbf{x}^{(s)}_i$ is determined by:
\begin{equation}
\label{eq:q}
q^{(s)}_{ij} =
\begin{cases}
1, & j = \arg\max\limits_{1 \le t \le k^*} \sum_{\theta=1}^{\Theta} f_{\theta} \mathbb{I} \left[ x^{(s)}_{i\theta} = c^{(s)}_{t\theta} \right], \\
0, & \text{otherwise}.
\end{cases}
\end{equation}
The computation of $f_{\theta}$ relies on the overall intra-feature similarity score $\mathfrak{f}_{\theta}$, formulated as:
\begin{equation}
\label{eq:intra_feature}
\mathfrak{f}_{\theta} = \sum_{i=1}^{n^{(s)}} \sum_{j=1}^{k^*} q^{(s)}_{ij} \mathbb{I} \left[ x^{(s)}_{i\theta} = c^{(s)}_{j\theta} \right],
\end{equation}
Accordingly, the normalized feature weight $f_{\theta}$ is computed by:
\begin{equation}
\label{eq:f}
f_{\theta} = \frac{\mathfrak{f}_{\theta}}{\sum_{t = 1}^{\Theta}\mathfrak{f}_{t}}. 
\end{equation}
To optimize the objective function in Eq.~\eqref{eq:objective}, Fed-HIRE jointly optimizes the partition matrix $\mathbf{Q}^{(s)}$ and the feature weights $\mathbf{f}$ by alternately updating one variable while holding the other fixed.

\section{Experiment}

\noindent\textit{\textbf{Experiment Settings.}}
\textbf{Four experiments} are conducted to validate our framework: 1) clustering performance evaluation, 2) ablation study, 3) multi-granular hierarchy effectiveness evaluation, and 4) efficiency analysis. 

\textbf{Counterparts.} 
Four SOTA counterparts are evaluated in our experiments, including Fed-SC~\cite{10184550}, NN-FC~\cite{10634982}, SFK~\cite{10931177}, and FKDC~\cite{10931177}. The hyperparameters for these baselines are configured in strict accordance with their respective original papers. 

\textbf{Benchmark.} 
Six datasets, comprising four real-world datasets and two synthetic datasets, are employed for empirical validation. The four real-world datasets are sourced from the UCI Machine Learning Repository~\cite{dua2017uci}. Detailed characteristics and statistics of all datasets are summarized in Table~\ref{tbl:dataset}.

\textbf{Non-IID Simulation with Incomplete Clusterlets.}
To simulate a non-IID setting with incomplete clusterlets, we first run $k$-means on the original dataset and divide all samples into $\alpha \times L$ clusterlets, where $L$ denotes the number of clients and $\alpha$ controls the average number of clusterlets assigned to each client. In our experiments, we set $\alpha=3$.
For the $l$-th client, we randomly assign $z_l$ clusterlets, where $z_l \in \{\alpha-1, \alpha, \alpha+1\}$. The three candidate values are chosen with equal probability, subject to the constraint $\sum_{l=1}^{L} z_l = \alpha L$. The clusterlets are then assigned to clients uniformly at random without replacement. The local dataset of the $l$-th client is constructed as the union of its assigned clusterlets.
This process fragments each global cluster across multiple clients, thereby simulating a practical non-IID setting with incomplete clusters.

\textbf{Implementation Details.}
For the proposed Fed-HIRE, we set the learning rate to $\eta=0.0001$. At the beginning of each optimization stage in both ACD and MHR, the winning-frequency counter $g_j$ and the auxiliary variable $\tau_j$ are initialized to 1 for every active candidate clusterlet, i.e., $g_j=\tau_j=1$ for all active $j$. The initial numbers of clusters are set to $k_0=0.5n^{(l)}$ in the local ACD stage and $k_0=0.5n^{(s)}$ in the global MHR stage. For all evaluated methods, the target number of clusters $k^*$ is set as reported in Table~\ref{tbl:dataset}. For the VCC mechanism, the maximum sampling cardinality is set to $\delta=3$ in all experiments. Unless otherwise stated, the number of clients is fixed to $L=10$.

\begin{table}[!t]
\centering
\caption{Statistics of the experimental datasets.}
\label{tbl:dataset}
\resizebox{\columnwidth}{!}{
\begin{tabular}{c c c c|c c c c}
\toprule
\textbf{Dataset} & \textbf{$n$} & \textbf{$d$} & \textbf{$k^*$} & \textbf{Dataset} & \textbf{$n$} & \textbf{$d$} & \textbf{$k^*$} \\
\midrule
Iris(IR)            & 150    & 4    & 3  & Accent(AC)          & 329    & 12        & 6  \\
Yeast(YE)           & 1484   & 8    & 10 & Wine Quality(WI)    & 6497   & 11        & 7  \\
Synthetic N(SN)     & 100000 & 20   & 10 & Synthetic D(SD)     & 1000   & 100000    & 10 \\
\bottomrule
\end{tabular}
}
\end{table}

\begin{table}[!t]
    \centering
    \caption{Federated Clustering Performance w.r.t. Purity (upper) and Ablation Study (lower). The ``Ave.'' column denotes the average ranks and the average proportion of degraded performance. The notations $\mathcal{F}$ and $\mathcal{M}$ represent the key components ACD and MHR, respectively, of Fed-HIRE.
    }
    \label{tbl:performance}
    \resizebox{\columnwidth}{!}{
    \begin{tabular}{c|c c c c|c|c}
    \toprule
    \textbf{Baseline} & \textbf{IR} & \textbf{AC} & \textbf{YE} & \textbf{WI} & \multicolumn{2}{c}{\textbf{Ave.}} \\
    \midrule
    \text{Fed-SC} & \text{0.740±0.00} & \text{0.513±0.01} & \text{0.447±0.01} & \text{0.453±0.01} & \text{2.5} & \multirow{5}{*}{rank} \\ 
    \text{NN-FC} & \text{0.490±0.14} & \text{0.5018±0.00} & \text{0.3280±0.02} & \text{0.4388±0.00} & \text{4.5} & \\
    \text{SFK} & \text{0.6787±0.06} & \text{0.5015±0.00} & \text{0.3482±0.02} & \text{0.4609±0.02} & \text{3.5} & \\
    \text{FKDC} & \text{0.6573±0.08} & \text{0.5015±0.00} & \text{0.3673±0.03} & \text{0.4705±0.01} & \text{3.5} & \\
    \text{Fed-HIRE (Full)} & \textbf{0.7867±0.06} & \textbf{0.5805±0.02} & \textbf{0.4650±0.02} & \textbf{0.4734±0.01} & \textbf{1} \\
    \midrule
    \text{w/o $\mathcal{F} + \mathcal{M}$} & \text{0.6147±0.05} & \text{0.5015±0.03} & \text{0.4012±0.06} & \text{0.4531±0.06} & $-$\textcolor{green}{13.37\%} & \multirow{3}{*}{\begin{tabular}[c]{@{}c@{}} $\Delta$ \\ vs. \\ Full \end{tabular}} \\
    \text{w/o $\mathcal{F}$} & \text{0.6733±0.10} & \text{0.5289±0.09} & \text{0.4263±0.01} & \text{0.4628±0.04} & $-$\textcolor{green}{8.47\%} & \\
    \text{w/o $\mathcal{M}$} & \text{0.6667±0.08} & \text{0.5258±0.05} & \text{0.4198±0.04} & \text{0.4696±0.03} & $-$\textcolor{green}{8.80\%} & \\
    \bottomrule
    \end{tabular}
    }
\end{table}

\noindent\textit{\textbf{Clustering Performance Evaluation.}}
The top panel of Table~\ref{tbl:performance} compares Fed-HIRE with state-of-the-art baselines in terms of Purity~\cite{schutze2008introduction}, where a higher value indicates better clustering performance. Each result is averaged over 10 independent runs, and the best result on each dataset is highlighted in \textbf{boldface}. The ``Ave.'' column reports the average rank of each method across all datasets. Fed-HIRE achieves the highest Purity on all four datasets. The improvements are particularly pronounced on IR and AC, while Fed-HIRE retains a smaller but consistent advantage on the more competitive WI dataset. Moreover, Fed-HIRE attains the optimal average rank of 1.0, substantially outperforming its counterparts. These results demonstrate the consistent effectiveness of Fed-HIRE across the evaluated datasets.

\noindent\textit{\textbf{Ablation Study.}}
To evaluate the individual and joint contributions of the client-side ACD and server-side MHR modules, we conduct an ablation study, with the results summarized in Table~\ref{tbl:performance} (lower). The notation ``w/o'' specifies the component omitted from the framework, where $\mathcal{F}$ and $\mathcal{M}$ represent the exclusion of ACD and MHR, respectively. The ``Ave.'' column reports the average percentage of performance degradation across the ablation variants. Removing either module leads to consistent performance drops compared to the full framework, confirming that both components contribute to the overall gains. Notably, the full Fed-HIRE framework achieves the best performance, demonstrating that ACD and MHR are complementary.

\begin{figure}[!t]
    \centering
    \begin{subfigure}[b]{0.49\linewidth}
        \centering
        \includegraphics[width=\textwidth]{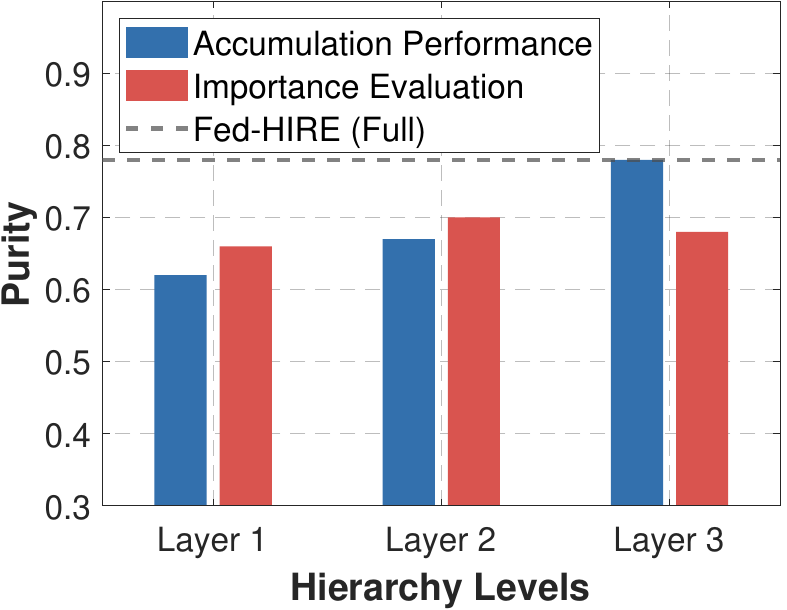} 
        \caption{Dataset: IR}
        \label{fig:sub_a}
    \end{subfigure}
    \begin{subfigure}[b]{0.49\linewidth}
        \centering
        \includegraphics[width=\textwidth]{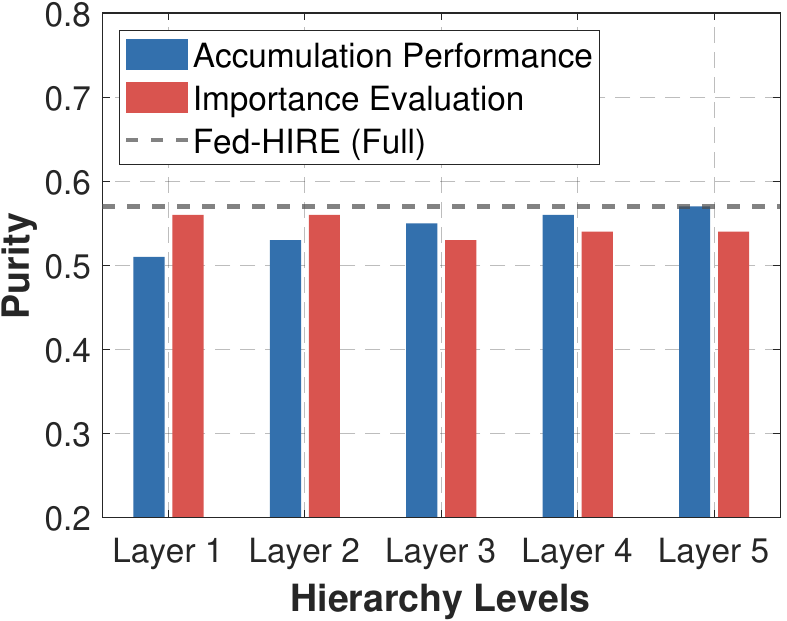} 
        \caption{Dataset: AC}
        \label{fig:sub_b}
    \end{subfigure}
    \begin{subfigure}[b]{0.49\linewidth}
        \centering
        \includegraphics[width=\textwidth]{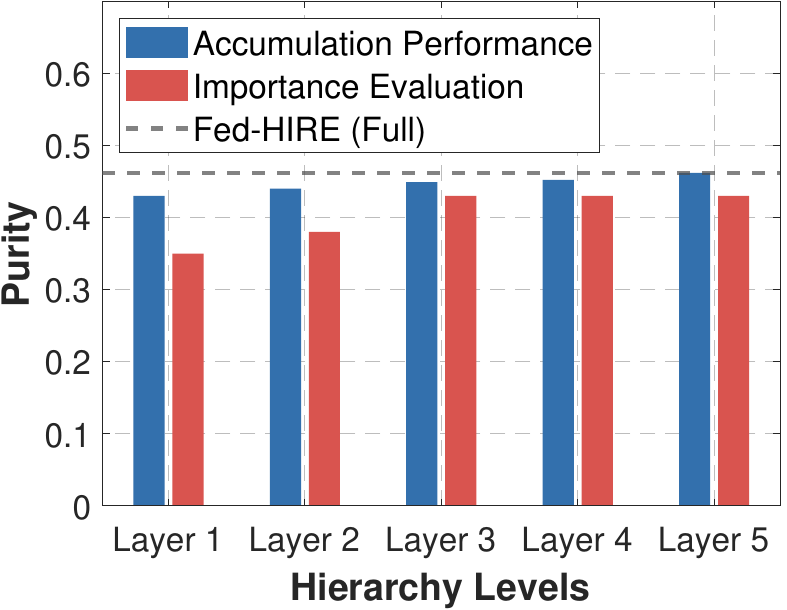} 
        \caption{Dataset: YE}
        \label{fig:sub_c}
    \end{subfigure}
    \begin{subfigure}[b]{0.49\linewidth}
        \centering
        \includegraphics[width=\textwidth]{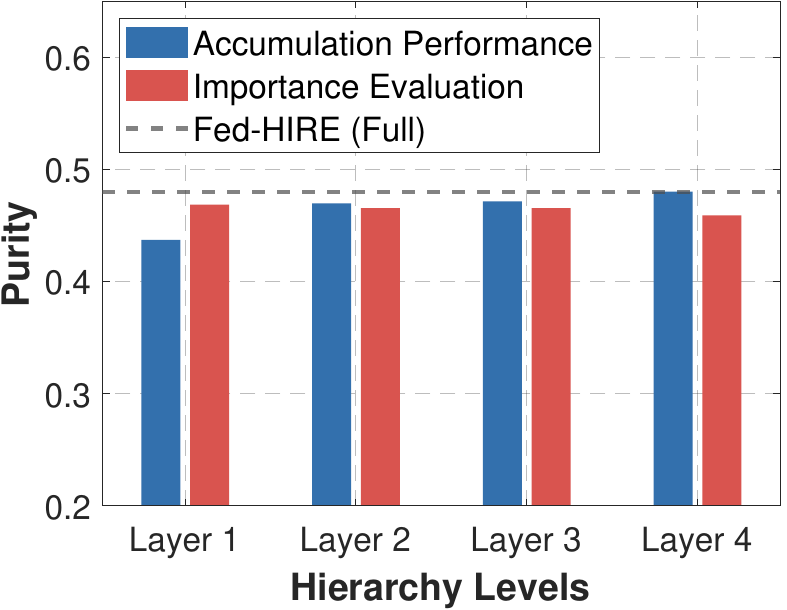} 
        \caption{Dataset: WI}
        \label{fig:sub_d}
    \end{subfigure}
    \caption{Effectiveness of the proposed multi-granular hierarchy on four representative datasets.}
    \label{fig:hierarchy}
\end{figure}

\noindent\textit{\textbf{Multi-granular Hierarchy Effectiveness Evaluation.}}
Fig.~\ref{fig:hierarchy} evaluates the effect of the proposed multi-granular hierarchy on four datasets. The dashed line reports the performance of the full Fed-HIRE framework. We first examine the effect of gradually adding hierarchical levels to the clustering process, starting from the finest level and ending with the full hierarchy (blue bars). The clustering performance generally improves as more levels are included, suggesting that different granularities provide complementary distributional information that is not captured by a single partition. This helps Fed-HIRE recover the global clustering structure more reliably.
We further assess the contribution of each level by removing one level at a time from the full hierarchy (red bars). On YE, removing the first level leads to a larger performance drop, which indicates that fine-grained clusterlets are important for preserving local distribution details. On IR, the contribution of the coarser third level suggests that coarse semantics help align fragmented structures. The patterns on AC and WI also vary across levels, showing that the most useful granularity depends on the dataset. Overall, these results confirm that each hierarchical level provides distinct information for aligning the global distribution.

\begin{figure}[!t]
  \centering
  \includegraphics[width=0.49\linewidth]{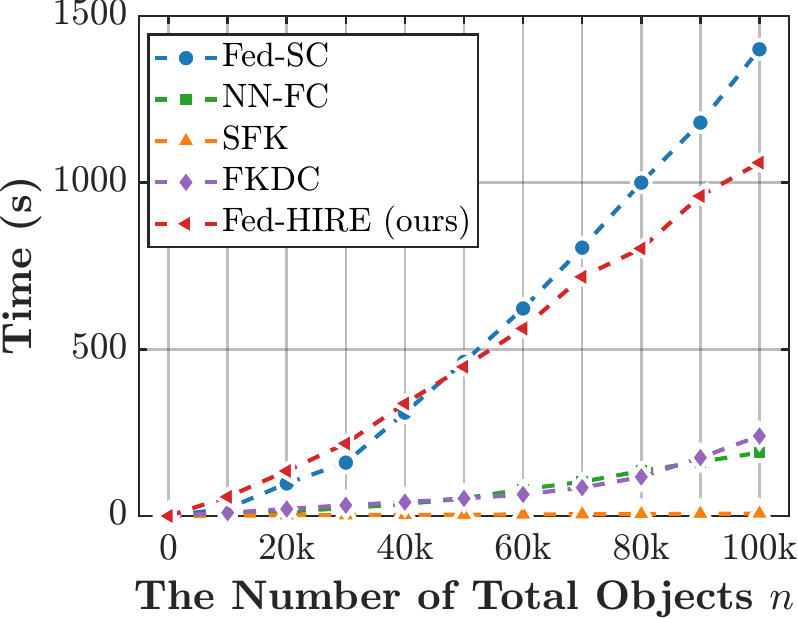}
  \includegraphics[width=0.49\linewidth]{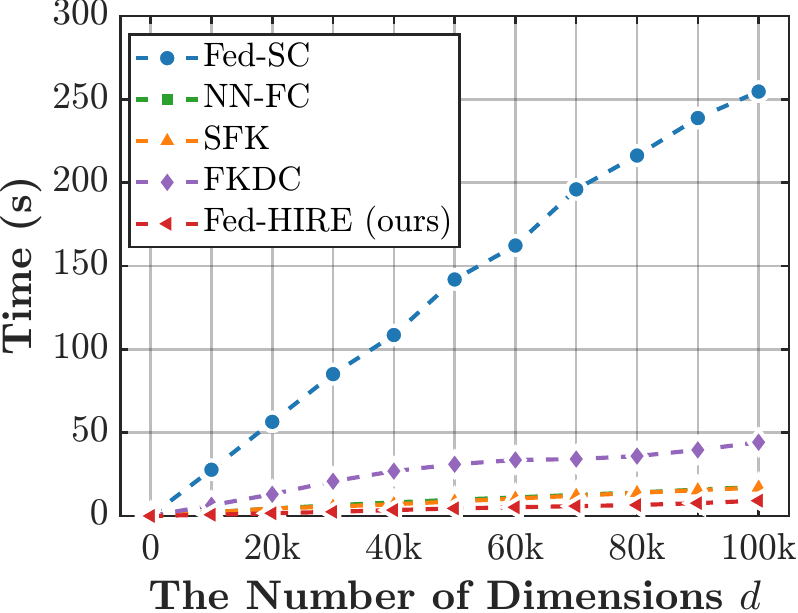}
  \caption{Efficiency evaluation of Fed-HIRE and the baselines on large synthetic datasets with respect to the total number of objects $n = \sum_{l=1}^{L} n^{(l)}$ on SN and the dimensionality $d$ on SD.}
  \label{fig:efficiency}
\end{figure}

\noindent\textit{\textbf{Efficiency Evaluation.}}
Fig.~\ref{fig:efficiency} compares the execution times of Fed-HIRE and the baselines on the large synthetic datasets SN and SD while varying the total number of objects $n = \sum_{l=1}^{L} n^{(l)}$ and the dimensionality $d$, respectively. Fed-HIRE exhibits near-linear scalability with respect to both $n$ and $d$, demonstrating its applicability to large-scale, high-dimensional datasets. Although Fed-HIRE requires more execution time than some baselines, this computational overhead represents a reasonable trade-off for discovering multi-granular hierarchical structures.

\section{Concluding Remarks}
This paper introduces Fed-HIRE, a novel one-shot federated clustering framework that tackles the challenging problem of unsupervised distribution aggregation when global clusters are highly fragmented across heterogeneous clients. By bridging client-side fine-grained distribution exploration with server-side multi-granular aggregation, Fed-HIRE effectively reconstructs coherent global hierarchies, circumventing the prohibitive communication overhead and privacy vulnerabilities typical of iterative methods. Beyond establishing a robust empirical baseline, this work provides a highly scalable paradigm for unsupervised knowledge discovery in complex, open-environment distributed networks. As Fed-HIRE currently focuses on tabular data, a compelling avenue for future research is extending its multi-granular structural alignment to accommodate more complex streaming or unstructured modalities.

\begin{acks}
This work was supported in part by the National Natural Science Foundation of China (NSFC) under Grants: 62476063, 62376233, and 62306181; the Natural Science Foundation of Guangdong Province under Grant: 2025A1515011293; the General Research Fund of RGC under Grants: 12202025 and 12201323; the Guangdong and Hong Kong Universities ``1+1+1'' Joint Research Collaboration Scheme under Grant: 2025A0505000004; and the Initiation Grant for Faculty Niche Research Areas (IG-FNRA) of HKBU with the grant: RC-FNRA-IG/23-24/SCI/02.
\end{acks}

\section*{GenAI Usage Disclosure}
Generative AI (GenAI) tools were used solely for language polishing and grammar improvements in this paper. 

\bibliographystyle{ACM-Reference-Format}
\bibliography{reference}

\end{document}